\documentclass[10pt,conference]{IEEEtran}
\IEEEoverridecommandlockouts

\usepackage{cite}
\usepackage{algorithm}
\usepackage{algpseudocode}
\usepackage{graphicx}
\usepackage{textcomp}
\usepackage{xcolor}
\usepackage{dsfont}

\def\BibTeX{{\rm B\kern-.05em{\sc i\kern-.025em b}\kern-.08em
T\kern-.1667em\lower.7ex\hbox{E}\kern-.125emX}}
\usepackage{afterpage}
\usepackage{mathtools}
\usepackage{cuted}
\usepackage{subfigure}
\usepackage{array}
\usepackage{epsf}
\usepackage{times}
\usepackage{epsfig}
\usepackage{epstopdf}
\usepackage{mathtools}
\usepackage{amsthm}
\usepackage{amssymb}
\usepackage{comment}
\usepackage{siunitx}
\usepackage{lipsum}
\usepackage{kantlipsum}
\usepackage{dblfloatfix}
\usepackage{adjustbox}
\usepackage[font=small]{caption}
\usepackage{subcaption}
\usepackage{amsmath}
\usepackage{amsfonts}
\usepackage{mathtools}
\usepackage{amscd}
\usepackage{bm}
\usepackage{subfigure}

\usepackage{tikz}
\usepackage{pgfplotstable}
\usepackage{rotate}

\definecolor{rubblue}{cmyk}{1,0.5,0,0.6}
\definecolor{rubgreen}{cmyk}{0.5,0,1,0}
\definecolor{rubgray}{cmyk}{0.03,0.03,0.03,0.1}

\usepgfplotslibrary{units}

\usetikzlibrary{%
patterns,%
calc,%
fit,%
arrows,%
plotmarks,%
shadows,%
chains,%
shapes%
}

\tikzset{>=latex'} 
\tikzstyle{every picture}+=[remember picture] 
\pgfdeclarelayer{background}
\pgfdeclarelayer{foreground}
\pgfsetlayers{background,main,foreground}

\tikzstyle{blueblock}=[draw=rubblue, rectangle, thick, drop shadow, minimum width=20mm, minimum height=8mm,fill=rubblue!20, text width=20mm, text centered]
\tikzstyle{bluebox}=[draw=rubblue, rectangle, thick, drop shadow, minimum width=8mm, minimum height=8mm,fill=rubblue!20, text width=8mm, text centered]
\tikzstyle{greenblock}=[draw=rubgreen, rectangle, thick, drop shadow, minimum width=20mm, minimum height=8mm,fill=rubgreen!20, text width=20mm, text centered]
\tikzstyle{dot} = [draw, circle, minimum size=0.2pt,scale=0.3,fill=black,black]
\tikzstyle{smalldot} = [draw, circle, minimum size=0.1pt,scale=0.2,fill=black,black]
\tikzstyle{reddot}  =[draw,circle,minimum size=0.2pt,scale=0.8,fill=red,thin]
\tikzstyle{greendot}  =[draw,circle,minimum size=0.2pt,scale=0.8,fill=Green,thin]
\tikzstyle{bluedot}  =[draw,circle,minimum size=0.2pt,scale=0.8,fill=blue,thin]
\tikzstyle{whitedot}=[draw,circle,minimum size=0.2pt,scale=0.8,fill=white,thin]
\tikzstyle{blackdot} = [draw, circle, minimum size=0.2pt,scale=0.7,fill=black,black]
\tikzstyle{sum} = [drop shadow, draw=rubblue, thick, fill=rubblue!20, circle]
\tikzstyle{relay} = [blueblock, minimum width=5mm, minimum height=20mm, text width=5mm, rounded corners=2pt]
\tikzstyle{relay2} = [blueblock, minimum width=5mm, minimum height=15mm, text width=5mm, rounded corners=2pt]
\tikzstyle{relay3} = [blueblock, minimum width=5mm, minimum height=25mm, text width=5mm, rounded corners=2pt]
\tikzstyle{relay4} = [blueblock, minimum width=5mm, minimum height=10mm, text width=5mm, rounded corners=2pt]
\tikzstyle{relay5} = [blueblock, minimum width=5mm, minimum height=50mm, text width=5mm, rounded corners=2pt]
\tikzstyle{relay6} = [blueblock, minimum width=5mm, minimum height=5mm, text width=5mm, rounded corners=2pt]
\tikzstyle{circgreen} = [draw, circle, inner sep=2pt, fill=rubgreen, drop shadow, thick]
\tikzstyle{circwhite} = [draw, circle, inner sep=2pt, fill=white, drop shadow, thick]
\tikzstyle{circdashed} = [draw, dashed, circle, inner sep=2pt, fill=rubgray, drop shadow, thick]
\tikzstyle{vertbox} = [rectangle, draw=rubblue, thick, rotate=90, text centered, minimum width=16.5mm, minimum height=8mm, text width=16.5mm, inner sep=0pt, fill=rubblue!20, drop shadow]
\tikzstyle{vertboxb} = [rectangle, draw=rubblue, thick, rotate=90, text centered, minimum width=16.5mm, minimum height=8mm, text width=16.5mm, fill=rubblue!20, drop shadow]
\tikzstyle{vertboxshort} = [rectangle, draw=rubblue, thick, rotate=90, text centered, minimum width=10mm, minimum height=8mm, text width=10mm, inner sep=0pt, fill=rubblue!20, drop shadow]
\tikzstyle{smalldotgreen} = [draw=rubgreen, circle, minimum size=0.2pt,scale=0.8,fill=rubgreen!20]
\tikzstyle{antenna} = [regular polygon, regular polygon sides=3, draw, shape border rotate=180, minimum size=0.2pt, scale=0.3]

\tikzstyle{poly} = [regular polygon, regular polygon sides=6, shape aspect=0.5, minimum width=1.5cm, minimum height=0.35cm, draw, dashed]

\definecolor{cff9e00}{RGB}{255,158,0}
\definecolor{c4fff00}{RGB}{79,255,0}
\definecolor{cff0012}{RGB}{255,0,18}
\definecolor{c00c5ff}{RGB}{0,197,255}
\definecolor{c046f00}{RGB}{4,111,0}
\definecolor{c004b9d}{RGB}{0,75,157}

\newlength{\mylen}
\usetikzlibrary{petri}
\usetikzlibrary{shapes}
\usetikzlibrary{positioning,arrows,patterns}
\usepackage{scalefnt}
\usetikzlibrary{calc,decorations.markings}
\pgfplotsset{compat=1.10}
\usetikzlibrary{arrows.meta}
\usepackage[columnwise,switch,mathlines]{lineno} 
\usepackage{pgfplots}
\pgfplotsset{compat=newest}
\usepgfplotslibrary{groupplots}

\IEEEoverridecommandlockouts
\newtheorem{theorem}{Theorem}

\newtheorem{defn}{\textbf{Definition}}

\newtheorem{remark}{Remark}

\begin{document}
\title{Decentralized Fairness Aware Multi Task \\
Federated Learning for VR Network}
\vspace{-0.7cm}
\author{\IEEEauthorblockN{Krishnendu S. Tharakan, \IEEEmembership{Member, IEEE}, Carlo Fischione, \IEEEmembership{Fellow, IEEE}}\\ \vspace{-0.5cm}\IEEEauthorblockA{ School of Electrical Engineering and Computer Science, KTH Royal Institute of Technology, Stockholm, Sweden\\ Email: $\{$tharakan, carlofi$\}@$kth.se}\\ }
\vspace{-1.5cm}
\setlength{\belowdisplayskip}{2pt}
\maketitle
\vspace{-5.5cm}
\begin{abstract}
Wireless connectivity promises to unshackle virtual reality (VR) experiences, allowing users to engage from anywhere, anytime. However, delivering seamless, high-quality, real-time VR video wirelessly is challenging due to the stringent quality of experience requirements, low latency constraints, and limited VR device capabilities. This paper addresses these challenges by introducing a novel decentralized multi task fair federated learning (DMTFL) based caching that caches and prefetches each VR user's field of view (FOV) at base stations (BSs) based on the caching strategies tailored to each BS. In federated learning (FL) in its naive form, often biases toward certain users, and a single global model fails to capture the statistical heterogeneity across users and BSs. In contrast, the proposed DMTFL algorithm personalizes content delivery by learning individual caching models at each BS. These models are further optimized to perform well under any target distribution, while providing theoretical guarantees via Rademacher complexity and a probably approximately correct (PAC) bound on the loss. Using a realistic VR head-tracking dataset, our simulations demonstrate the superiority of our proposed DMTFL algorithm compared to baseline algorithms.
\end{abstract}
\IEEEpeerreviewmaketitle
\vspace{-0.45cm}
\section{Introduction}
Virtual reality (VR) technology is poised to transform the way users engage with their surroundings, with the global VR market expected to exceed $435$ billion USD by $2030$~\cite{grandview23}. Realizing this potential, however, hinges on addressing a major challenge, which is ensuring reliable connectivity and high-quality user experiences over inherently unstable wireless channels~\cite{fenghe20}. 


VR applications heavily depend on $360^{\circ}$ immersive video technology to create compelling user experiences~\cite{chen18}. While current video coding techniques often tailor streaming to user attention and field of view (FOV)~\cite{Zink2019}, delivering real-time, tile-based FOV content remains a major challenge. The steps involved in detecting, retrieving, and transmitting the appropriate tiles for a user's FOV can introduce considerable latency, an issue that worsens as the number of users grows. This latency can critically degrade the user experience, making it hard to meet stringent delay requirements. 
Among them, proactive caching of VR content at the network edge stands out as a particularly promising solution to improve delivery efficiency and reduce latency~\cite{guo24, tharakan25}.

Much of the existing research on VR content delivery concentrates on enhancing the throughput of wireless VR networks through advanced wireless resource allocation methods. For instance, the work in~\cite{chen22}, proposes an iterative algorithm for optimizing wireless transmission in multiplayer interactive VR gaming frameworks using mobile edge computing (MEC), which employs a truncated first-order Taylor expansion of the objective function, which is progressively refined through iterative updates. Similarly, \cite{gupta23} introduces a two-phase method for associating VR users with mmWave access points, the first phase applies a graph-theoretic assignment technique, followed by optimization using geometric programming. 

\vspace{-0.05cm}

To the best of our knowledge, leveraging a decentralized multi-task fair federated learning (DMTFL) framework for VR content delivery while explicitly accounting for the users’ FOV requests, and promoting fairness among users has not yet been explored within the wireless communications domain. Motivated by this gap, we tackle the caching problem through a decentralized learning framework, without relying on prior statistical assumptions about the request sequence. 
This paper proposes a novel wireless VR network architecture empowered by MEC. Our proposed DMTFL algorithm enables personalized caching strategies at each BS, ensuring equitable service quality and robustness even under non-independent and identically distributed (non-iid) user request patterns. 

To this end, we present a unified framework with rigorous theoretical guarantees and introduce a DMTFL based caching algorithm. The fairness aspect ensures that caching strategies are not biased toward particular BSs or user distributions, enabling equitable service quality across the network. Meanwhile, the multi-task aspect allows the proposed DMTFL algorithm to learn separate caching models for each BS, thereby providing personalized and locally optimized strategies tailored to the data and user demands at each BS. Further, to theoretically justify the performance of the proposed decentralized caching strategies, we incorporate tools from statistical learning theory, specifically the probably approximately correct (PAC) framework. 
\vspace{-0.05cm}

Furthermore, we use Rademacher complexity as a measure of function class capacity, quantifying the ability of our learning algorithm to generalize beyond the observed samples. This approach enables us to formally establish a bound (Theorem~\ref{thm:mainresult_cachingvr}) that ensures the proposed algorithm not only performs well on the training data but also offers reliable guarantees on future requests, which is crucial for robust VR content delivery. Simulation results demonstrate that the proposed algorithm can substantially improve communication performance.

\vspace{-0.2cm}
\section{System Model and Problem Formulation} \label{sec:sys_model}
\begin{figure}[t!]
    \centering
\begin{tikzpicture}
   \node (system) at (0,0)
{\includegraphics[width=4cm]{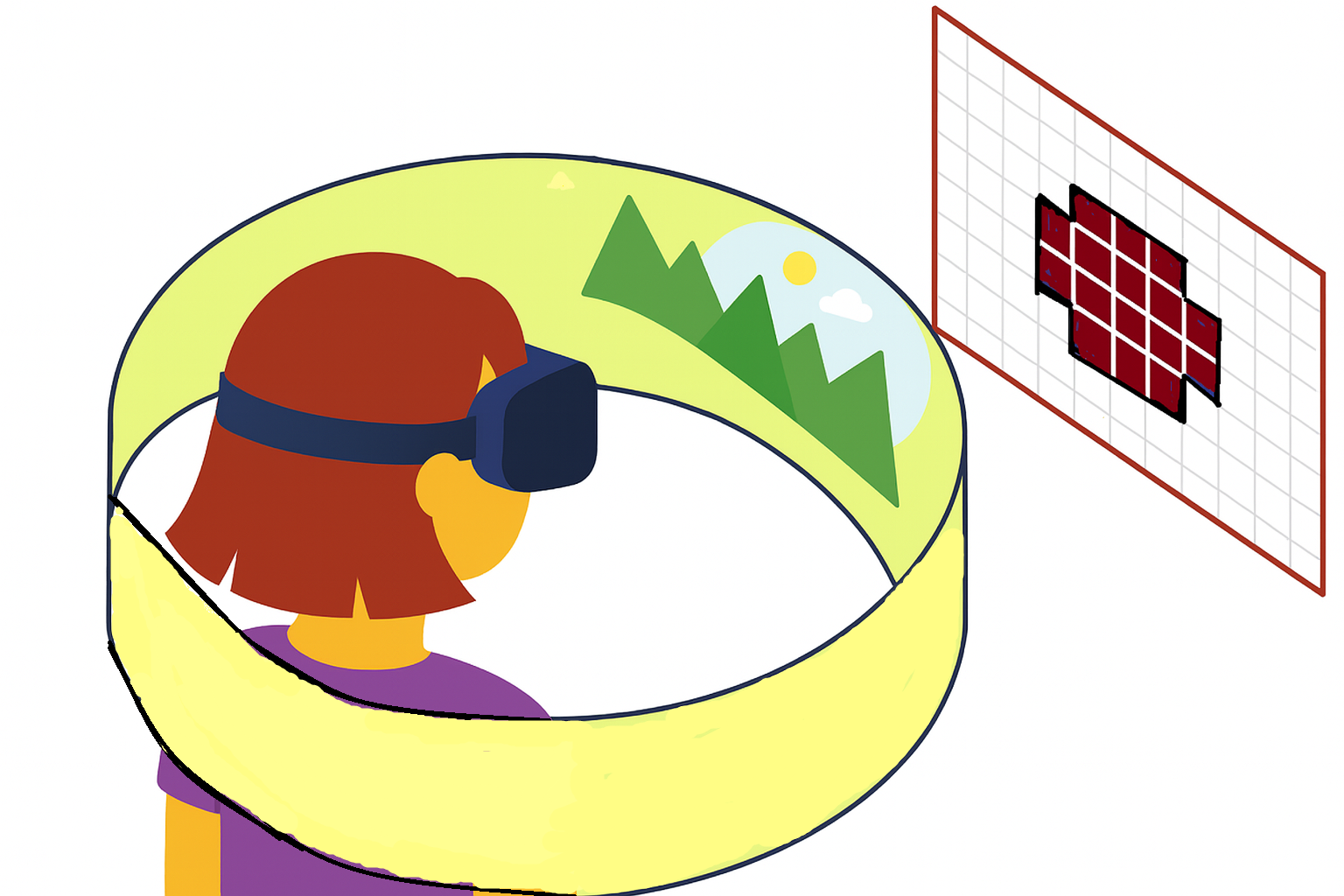}};
    \node [] (C) at (3.5,1.4) {Tiles in FOV};
   \node (D) at (3.1,1.35) {};
\node (E) at (1.6,0.8) {};
\draw[->, bend left] (D) to (E);
\node [] (C) at (-1.8,1.3) {FOV frame area};
\node (A) at (-1.1,1.35) {};
\node (B) at (0.6,0.5) {};
\draw[->, bend left] (A) to (B);
\node [] (C) at (2.8,-0.94) {Tile grid in 2D};
\node [] (C) at (3,-1.3) {EQR projection};
\node (F) at (1.9,-1.4) {};
\node (G) at (1.8,-0.3) {};
\draw[->, bend left] (F) to (G);
\end{tikzpicture}
\caption{Tiled FOV representation based on a user’s FOV within the equirectangular (EQR) projection of a $360^{\circ}$ VR scene}
 \label{fig:sys_model1}  
\end{figure}
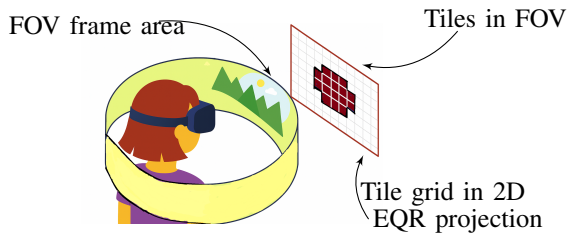

We consider $B$ BSs denoted by the set $\mathcal{B} = \{1,2,\ldots,b,\ldots,B\}$ and $U$ users denoted by the set $\mathcal{U} = \{1,2,\ldots,i,\ldots,U\}$. In this decentralized setting, BSs communicate and exchange models through limited-capacity links.  Following a similar approach to \cite{lungaro18}, we assume that only a portion of the full $360^{\circ}$ VR video is requested by a user referred to as the FOV. To facilitate processing, the $360^{\circ}$ video is projected onto a two-dimensional plane and divided into $F=N\times P$ tiles. The user requests a subset of tiles from the set $\mathcal{F} =\{1,2,\ldots,f,\ldots,F\}$, as illustrated in Fig.~\ref{fig:sys_model1}. 

\noindent We denote the set of users connected to the $b$-th BS as $\mathcal{U}_b$. Let the feature vector for the $b$-th BS for the $f$-th FOV (eg. past demand, FoV history, etc.) be denoted as $x_{b,f}$. The total demand for the $f$-th FOV at the $b$-th BS is given by $y_{b,f}$, where $y_{b, f}$ serves as the target label. To evaluate the VR video quality, we use the mean squared error (MSE) metric. Let $c_{b,f}$ be the binary indicator representing the $f$-th cached FOV at the $b$-th BS, and define the MSE for the $f$-th FOV from the $b$-th BS as ${e}_{b,f} = (c_{b,f} - y_{b,f})^2.$ Furthermore, to account for data heterogeneity, the data across devices is assumed to be non-iid, and are drawn from an underlying distribution 
$\mathcal{D}_b$ associated with the $b$-th BS, with each $\mathcal{D}_b =  \big((x_{b,i}, y_{b,i}), \ldots, (x_{b,m_b}, y_{b,m_b})\big)$ of size $m_b$, which which captures the local data at the $b$-th BS, including user FOV requests. Let $\bm{\mathcal{D}} = \{\mathcal{D}_1, \ldots, \mathcal{D}_B \}$ represent the set of all distributions of the BSs. 

In this work, we propose a novel algorithm, which we call DMTFL, where each BS learns a personalized caching vector $\bm{\phi_b}$ by minimizing its local empirical loss while also incorporating information from neighboring BSs to promote fairness and avoid bias toward any particular BS. Formally, each BS employ a DMTFL based caching strategy given by $\bm{\phi}_b$, where $\bm{\phi}_{b} = [\phi_{b,1}, \phi_{b,2},\ldots,\phi_{b,F} ]$ denotes the caching strategy employed at the $b$-th BS for the $f$-th FOV. Let ${\Phi}= \{\bm{\phi}_1, \bm{\phi}_2, \ldots, \bm{\phi}_B\} \subset \mathbb{R}^{F \times B}$ be the set of all caching strategies. The effectiveness of the caching scheme is evaluated using the following loss function:
\vspace{-0.25cm}
\begin{equation}
\mathcal{L}_b(\bm{\phi}_b) = \mathbb{E}_{(x,y) \in \mathcal{D}_b}\sum_{f \in \mathcal{F}}{\phi_{b,f}}\log_{10}(e_{b,f}). 
\end{equation}
\vspace{-0.1cm}
\noindent For the ease of notation, we denote $\mathbb{E}_{(x,y) \in \mathcal{D}_b}\sum_{f \in \mathcal{F}}{\phi_{b,f}}\log_{10}(e_{b,f}) = \mathbb{E}_{(x,y) \in \mathcal{D}_b}\ell
(h_{\bm{\phi}}(\bm{x}),\bm{y})$, where $\ell$ is the loss function that combines the caching strategy and the mean squared error, and 
$h_{\bm{\phi}}(\bm{x})$ represents the caching policy predicted indicator vector for the given features.

Our primary goal is to evaluate the performance of the proposed algorithm by minimizing the loss function. The formal definition of the DMTFL algorithm is given later in Section~\ref{sec:algo_guarant}. It is important to recognize that employing a naive federated learning (FL) approach typically leads to learning a single, shared caching strategy across all BSs (i.e., $\bm{\phi}_b$ would be same across the BSs). While FL offers the advantage of leveraging local datasets and computation at edge BSs with minimal communication overhead, it also introduces specific challenges. In its standard form such as FedAvg~\cite{pmlr-mcmahan17a} or FedProx~\cite{li2020federated}, classical FL aims to train a single global model using data distributed across numerous, potentially millions of remote devices. This approach often overlooks the unique characteristics and contributions of individual BSs. When models are trained independently at different BSs with heterogeneous data distributions, the resulting statistical relationships may vary significantly. As a result, aggregating these into a single global model can lead to suboptimal performance and reduced inference accuracy. 

Another key problem in FL is that of fairness. In many practical cases, the resulting learning models can exhibit bias or lack fairness~\cite{mohri19a}. A key reason these concerns arise is that the target distribution for which the FL model is learned is often unspecified. To account for fairness, we consider an agnostic FL setting, where the caching strategy is optimized for any possible target distribution formed by a mixture of the BS distributions. Thus, to address these limitations, our work adopts a personalized FL (also known as multi-task FL) as well as fair FL (also known as agnostic FL) framework that learns separate models for each BS, explicitly accounting for any target distribution in the federated setting. Instead of tailoring the caching strategy to a particular distribution, which risks misalignment with the target, we formulate an agnostic and more robust objective. In this setting, the target distribution is modeled as an unknown mixture of the distributions $\mathcal{D}_b, b = 1,\ldots, B$, that is $\mathcal{D}_w = \sum_{b=1}^B w_b \mathcal{D}_b$ for some $w \in \Delta_B$, or any $w$ in a subset $\Lambda \in \Delta_B$, where $\Delta_B$ stands for the simplex over $B$. Given that the mixture weight $w$ is unknown, we must design a solution that is robust across all possible $w$ values within the simplex. This consideration naturally leads to the following optimization problem.
\vspace{-0.4cm}
\begin{eqnarray} \label{eq:the_problem}
\min_{\bm{\phi}_{1},\bm{\phi}_{2},\ldots, \bm{\phi}_{B}} \sup_{\mathbf{w} \in \Lambda}\bigg\{ \theta_{\Phi, \mathbf{w}} = \sum_{b=1}^B w_b\mathcal{L}_b(\bm{\phi}_{b}, \mathcal{D}_b) \bigg\},
\end{eqnarray}
\vspace{-0.12cm}
\noindent where the coefficients satisfy $\sum_b^B w_{b} = 1$, and $\mathbf{w} = (w_1, \ldots, w_B)$.
Note that to address the problem above, each BS first estimates its local loss $\mathcal{L}_b(\bm{\phi}_b)$ using the empirical average $\hat{\mathcal{L}}_b(\bm{\phi}_b, \mathcal{D}_b)$. This estimate relies entirely on the BS’s own data, which, if limited, can be quite noisy and lead to poor performance. A practical way to mitigate this issue is to also incorporate information from neighboring BSs, particularly when their data distributions are similar. In the ideal case of iid data across devices, simply averaging all estimates works well. However, in non-iid scenarios, treating all neighbors equally can degrade the accuracy of the loss estimate. A more refined approach is to assign carefully chosen weights to each BS’s data, allowing the loss estimate to better reflect the true local distribution. Once these optimal weights are determined, the overall model becomes significantly more robust and personalized. The caching strategy can be personalized by assigning distinct weights to each BS, as demonstrated in~\cite{smith17}. This results in the following optimization problem, which is solved in a distributed manner.
\vspace{-0.25cm}
\begin{equation}
\label{eq:the_main_problem}
\hspace{-0.1cm}\min_{\bm{\phi}_{1},\ldots, \bm{\phi}_{B}} \sup_{\mathbf{w} \in \Lambda}\bigg\{\widehat{\theta}_{\Phi, \mathbf{w}, \bm{\alpha}}(\bm{\mathcal{D}}) = \sum_{b=1}^B w_b\sum_{i=1}^B\alpha_{b,i}\mathcal{\hat{L}}_b(\bm{\phi}_{i}, \mathcal{D}_b) \bigg\},    
\end{equation}
\vspace{-0.05cm}
where the coefficients satisfy $\sum_i^B \alpha_{b,i} = 1$, and $\bm{\alpha} = \{\bm{\alpha}_1,\ldots,\bm{\alpha}_B \}$, and $\bm{\alpha}_b = \{{\alpha}_{b,1},\ldots,{\alpha}_{b,B} \}$. Our approach explicitly accounts for spatial heterogeneity across devices and further incorporates this heterogeneity into the weighting of each loss metric. The main contribution of this paper is to derive a PAC bound on the difference $\theta_{\Phi, \mathbf{w}} - \widehat{\theta}_{\Phi, \mathbf{w}, \bm{\alpha}}(\bm{\mathcal{S}})$.
The subsequent section delves into the details of this distributed \& fair multi-task FL based caching strategy.
\vspace{-0.2cm}
\section{DMTFL Algorithm} 
\label{sec:algo_guarant}
\begin{figure}[t!]
    \centering
\begin{tikzpicture}
   \node (system) at (0,0)
  {\includegraphics[width=4.2cm]{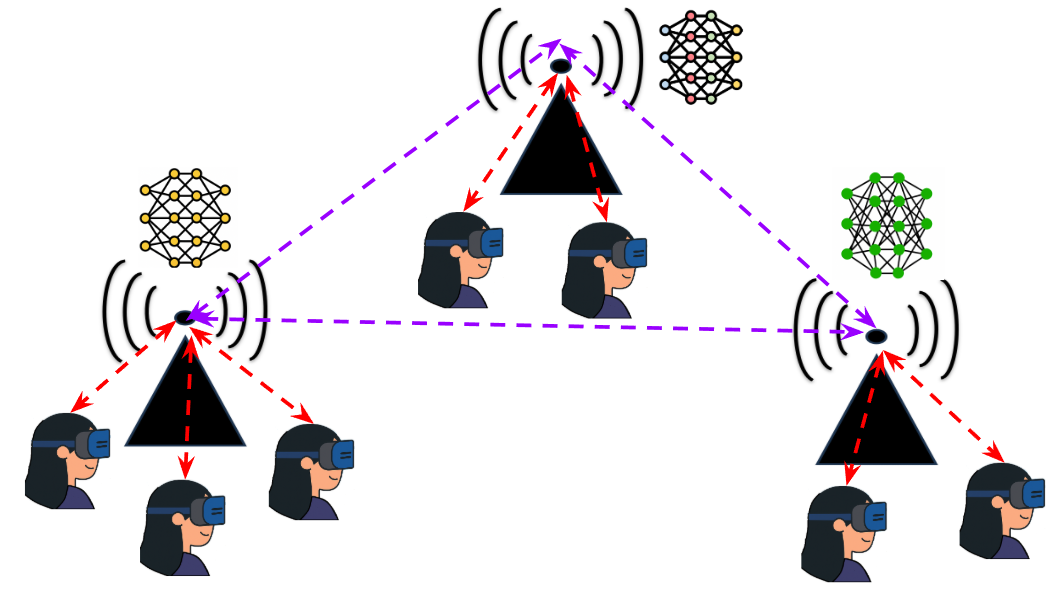}};
 \node [] (C) at (-3.0,-1.0) {VR Users};
\node [] (C) at (-0.8,1.1) {BS};
\end{tikzpicture}
\caption{Decentralized multi-task fair federated learning}
 \label{fig:sys_model2}  
\end{figure}

\vspace{-0.1cm}
\noindent Drawing on distributed and multi-task learning principles, we introduce a decentralized multi-task structure where each BS learns a personalized caching vector that accounts for both local user demands and cross-BS fairness considerations. As illustrated in Fig.~\ref{fig:sys_model2}, we adopt a fair and multi-task FL framework, where each BS learns its own personalized caching model using local data. By modeling the target distribution as an unknown mixture and allowing separate models per BS, this approach effectively captures statistical heterogeneity across BSs. Consequently, it ensures that caching decisions are both tailored to local data characteristics and robust to distributional shifts, thereby promoting fairness and improving overall system efficiency.
As depicted in Fig.~\ref{fig:sys_model2}, data collection across the network is non-iid, with variations in both data distribution and volume among BSs. In order to state the main result, we first introduce the following three key quantities: (i) Rademacher complexity, (ii) minimum $\epsilon$-cover, and (iii) discrepancy, each defined below.
\begin{defn} (Minimax weighted Rademacher complexity):
The Rademacher complexity for a given set of caching strategy ${\Phi}$ for a given $\mathbf{w} \in \Lambda$ is defined as
\vspace{-0.2cm}
\begin{eqnarray}
\mathcal{R}_{\mathbf{w}}({\Phi})\hspace{-0.1cm} =\hspace{-0.1cm} \underset{\substack{\mathcal{D}_1,., \mathcal{D}_B\\ \sigma}}
{\mathbb{E}} 
\bigg[ \underset{\substack{\bm{\phi}_b \in \bm{\Phi},\\ \alpha \in \Delta_B}}
{\sup} 
\sum_{b,i=1}^{B}\hspace{-0.1cm}\frac{w_b \alpha_{b,i}}{m_b}\hspace{-0.1cm}  
\sum_{j=1}^{m_b} \sigma_{bi,j} \, \ell \bigl( h_{\phi}(x_{b,j}), y_{b,j} \bigr) \bigg]\nonumber
\end{eqnarray}
where $\mathcal{D}_b = \big((x_{b,i}, y_{b,i}), \ldots, (x_{b,m_b}, y_{b,m_b})\big)$ is a sample of size $m_b$ and $\sigma_{bi,j} $ is a collection of Rademacher variables, that is uniformly distributed random variables taking values in $\{-1, 1\}$. max weighted Rademacher complexity is defined as $\mathcal{R}_{\Lambda}({\Phi}) = \max_{\mathbf{w} \in \Lambda} \mathcal{R}_{\mathbf{w}}({\Phi})$.
\end{defn}

\begin{defn} (Minimum $\epsilon$ cover):
The set $\{c_1, \ldots, c_d\}$ is said to be an $\epsilon$-cover of $\Lambda$ with respect to the $\ell_1$-distance if 
$\Lambda \subseteq \bigcup_{i=1}^{d} B(c_i, \epsilon),
$
where the $\ell_1$ ball is defined as 
$
B(c_i, \epsilon) := \{z \in \Lambda : \|z - c_i\|_1 < \epsilon\}.
$
The minimum $\epsilon$-cover, denoted by $\Lambda_\epsilon$, is an $\epsilon$-cover of $\Lambda$ with the smallest possible cardinality $d$.

\end{defn}

\begin{defn} (Discrepancy):
The discrepancy between two data distributions $\mathcal{D}_i$ and $\mathcal{D}_j$ corresponding to the BS $i$ and $j$, respectively, measured using the loss function $l: \mathcal{Y} \times \mathcal{Y} \rightarrow \mathbb{R}^+$ is defined as ${v}_{i,j} = \sup_{\bm{\phi}}| \mathcal{L}_i(\bm{\phi}) - \mathcal{L}_j(\bm{\phi})| $
\end{defn}
\vspace{-0.1cm}
The first term above represents the average loss at the $i$th BS, while the second term corresponds to the average loss at the $j$th BS. Their difference quantifies how distinct the two data distributions are with respect to the parameter $\bm{\phi}$. By maximizing over $\bm{\phi}$, we obtain the worst-case difference. When the data at the two BSs are iid, the discrepancy is zero. Conversely, if the data distributions differ significantly, the discrepancy will be large. A straightforward approach to incorporate this into an algorithm is to assign higher weights $\alpha$ when the discrepancy is small. In what follows, we present a theoretical result that offers a principled method for selecting these weights systematically. Specifically, the following theorem establishes a PAC bound on the difference between the true average loss in~\eqref{eq:the_problem} and its empirical estimate in~\eqref{eq:the_main_problem}, leveraging the definitions introduced above. By providing these theoretical guarantees, our proposed algorithm offers both practical benefits and a deeper understanding of its effectiveness in handling dynamic FOV request patterns.
\vspace{-0.25cm}
\subsection{ Theoretical Guarantees \& Proposed Algorithm}
\noindent In this subsection, we present a high probability bound on the performance of the proposed DMTFL algorithm. This analysis offers insights into selecting appropriate weights and optimizing the sequence of caching policies over time. 
\begin{theorem}
\label{thm:mainresult_cachingvr}
     (PAC Bound)
		Assuming the loss function is bounded, i.e., $\ell(a,b) < H, \forall a,b \in \mathcal{Y}$
        then for any $\epsilon > 0$, with probability at least $1- \delta$, $\delta >0 $, the following holds:        
\begin{equation}
    \label{eq:the_bound}
\theta_{\Phi, \mathbf{w}} \leq \widehat{\theta}_{\Phi, \mathbf{w}, \bm{\alpha}}(\bm{\mathcal{D}}) 
+ 2 \mathcal{R}_{\Lambda}({\Phi}) 
+ H \, \mathcal{P}(\mathbf{w}, \bm{\alpha}) 
+ H B\epsilon,
\end{equation}
where
$\mathcal{P}(\mathbf{w}, \bm{\alpha}) = 
\sqrt{
\frac{1}{2} 
\sum_{b=1}^{B}
\sum_{i=1}^{B}
\left(
\frac{w_b \alpha_{b,i}}{m_b}
\right)^2
\log \left( \frac{|\Lambda_{\epsilon}|}{\delta} \right)
}
\\
+ \frac{1}{H}
\sum_{b = 1}^{B}\sum_{i = 1}^{B} w_b \alpha_{b,i} v_{bi}$, 
$v_{bi}$ denotes the discrepancy, 
and $\Lambda_{\epsilon}$ is the minimum $\epsilon$-cover of $\Lambda$.
\end{theorem}
\begin{IEEEproof}\normalfont
    See Appendix \ref{appendix:proof_main_res_cachingvr}.
\end{IEEEproof} 
\begin{remark}
The covering parameter $\epsilon$ can be chosen moderately (e.g., $0.1$), since the PAC bound depends only logarithmically on $|\Lambda_\epsilon|$, ensuring practical feasibility even for large $B$.
\end{remark}
\vspace{-0.1cm}
\noindent Theorem \ref{thm:mainresult_cachingvr}
establishes a high-probability lower bound on the term $\theta_{\Phi, \mathbf{w}} - \widehat{\theta}_{\Phi, \mathbf{w}, \bm{\alpha}}(\bm{\mathcal{D}})$. Our goal is to optimize the caching strategy, by carefully selecting the coefficients $\alpha_{b,i}$, and $\bm{\phi}_b$, such that left hand term in Theorem~\ref{thm:mainresult_cachingvr} is minimized. To achieve this minimization, we solve the following problem as shown below
\begin{eqnarray}
\label{eq:min_prob}
\underset{\substack{{\bm{\phi}}_{1}, \ldots, \bm{\phi}_B\\ \bm{\alpha}}}
{\min} 
\bigg\{ \widehat{\Theta}_{\Phi, \mathbf{w}, \bm{\alpha}} \vspace{-0.5cm}&=&\vspace{-0.5cm}  \widehat{\theta}_{\Phi, \mathbf{w}, \bm{\alpha}}(\bm{\mathcal{D}}) \nonumber \\
&&
\vspace{-0.8cm}+ \sum_{b=1}^B\rho_b ||\bm{\phi}_b||_2
+ H \, \mathcal{P}(\mathbf{w}, \bm{\alpha}) 
\bigg\}.
\end{eqnarray}

\begin{algorithm}[h]
\captionof{algorithm}{DMTFL Algorithm}%
\label{alg:depe_fl_algo}
\begin{algorithmic}[1]
		\Procedure{Proposed DMTFL}{}
        \State \texttt{Initialize} $v_{b,i} = 1$, $\bm{\phi}_b^0$, $\alpha_{b,i}^0$, $\forall \, b,i$
        \For {$t = 1,2, \ldots, T$}
        \For {$b = 1,2, \ldots, B$}
        \For {$i = 1,2, \ldots, B$}
		\For {$n = 1,2, \ldots, N$} and $b \neq i$
            \State \text{Broadcast ${\bm{\phi}}_b^n$ to devices $b$ and $i$}
         \State \text{Obtain sub-gradients, and losses}
         \text{\hspace{2.5cm} from the BS $b$ and $i$ respectively}          
        \State \text{Update $v_{b,i}^n \vspace{-0.2cm} $}
        \begin{eqnarray}
		\hspace{2.1cm}v_{b,i}^n = |\widehat{\mathcal{L}}(\bm{\phi}_b^n, \mathcal{D}_b)- \widehat{\mathcal{L}}(\bm{\phi}_b^n, \mathcal{D}_i)| \nonumber
        \end{eqnarray}
        \State \text{Gradient ascent on $\bm{\phi}_b$}
        \begin{eqnarray}
		\hspace{2.1cm} \bm{\phi}_b^{n+1} = \bm{\phi}_b^{n} + \mu \partial_{\bm{\phi}}(\Delta_{b,i}(\bm{\phi}_b^n)) \nonumber
        \end{eqnarray}
        \EndFor
        \EndFor
        \State \text{Gradient descent on $\bm{\phi}_{b}$ }
        \begin{eqnarray}
		\vspace{-0.5cm}\bm{\phi}_{b}^{t+1} &=& {\bm{\phi}}_{b}^{t} - \eta \nabla_{\bm{\phi}_{b}}\widehat{\Theta}_{{\Phi}^{t}, \mathbf{w}, \bm{\alpha}^t} \nonumber
		\end{eqnarray}
        \State \text{Gradient descent with projection step on ${\bm{\alpha}}$ }
        \begin{eqnarray}
		\hspace{1.8cm}\alpha_{b}^{t+1}\ \gets\ \Pi_{\Delta_B}\!\Big(\alpha_{b}^{t}-\upsilon_t\,\nabla_{\alpha_b}\widehat{\Theta}_{\Phi^{t+1}, \mathbf{w}, \boldsymbol{\alpha}^{t}}\Big) \nonumber
		\end{eqnarray}
		\EndFor
        \State {Broadcast $\bm{\phi}_b$ to all BSs}
        \EndFor
        \State \texttt{Output final} $\{\bm{\phi}_b^{T}\}_{b=1}^B$ and $\{\alpha_{b,i}^{T}\}_{b,i=1}^B$
		\EndProcedure
  \end{algorithmic}
\end{algorithm}

\noindent Note that the above minimization problem requires knowledge of the discrepancy values, but since each BS only has access to its local data, these discrepancies must be estimated collaboratively in a distributed setup. Thus, the minimization problem described in~\eqref{eq:min_prob} must be addressed in a distributed manner, with minimal communication overhead between nodes. To enable this, BSs must approximate the discrepancies and use these estimates as proxies when designing the proposed DMTFL algorithm. As the discrepany terms are not directly observable, we utilize their estimates instead. To obtain these estimates, we employ a distributed gradient ascent algorithm, the steps of which are summarized in Algorithm~\ref{alg:depe_fl_algo}. It is noteworthy to mention that the described implementation of DMTFL (Algorithm~\ref{alg:depe_fl_algo}), provides the valuable flexibility to compute the estimates of individual caching strategies in a distributed manner across the network, with computations performed locally at each BS. 

\noindent From the definition of the discrepancy the proposed DMTFL algorithm essentially requires to solve 
\vspace{-0.3cm}
\begin{equation}
    {v}_{i,j} = \sup_{\bm{\phi}}| \mathcal{L}_i(\bm{\phi}) - \mathcal{L}_j(\bm{\phi})|, 
\end{equation}
for all the BSs. Since the true expected losses $\mathcal{L}_i(\bm{\phi})$ and $\mathcal{L}_j(\bm{\phi})$ are not directly accessible, they are approximated using their estimates denoted by $\widehat{\mathcal{L}}_i(\bm{\phi})$ and $\widehat{\mathcal{L}}_j(\bm{\phi})$ respectively. A natural approach to tackling this maximization problem is to apply a distributed gradient ascent algorithm. Algorithm~\ref{alg:depe_fl_algo} outlines the distributed implementation of this approach to estimate the discrepancies across the BSs. However, the presence of the absolute value in the discrepancy definition means that the gradient does not exist at all points. To overcome this issue, a generalized subgradient is used instead of the standard gradient as shown below. 
\vspace{-0.2cm}
\begin{eqnarray}
&&\hspace{-1.05cm}\partial_{\bm{\phi}}(\Delta_{b,i}(\bm{\phi}))\hspace{-0.1cm} 
=\nonumber \\
&&\bigl(\nabla \widehat{\mathcal{L}}(\bm{\phi}, \mathcal{D}_b)\hspace{-0.1cm}-\hspace{-0.1cm}\nabla \widehat{\mathcal{L}}(\bm{\phi}, \mathcal{D}_i)\bigr)\
\mathds{1}\hspace{-0.1cm} 
\{\widehat{\mathcal{L}}(\bm{\phi}, \mathcal{D}_b)\hspace{-0.1cm}>\hspace{-0.1cm}\widehat{\mathcal{L}}(\bm{\phi}, \mathcal{D}_i) \} \nonumber \\
&-&\hspace{-0.4cm}\bigl(\nabla \widehat{\mathcal{L}}(\bm{\phi}, \mathcal{D}_b) - \nabla \widehat{\mathcal{L}}(\bm{\phi}, \mathcal{D}_i)\bigr)\ \hspace{-0.2cm}
\mathds{1}\{ \widehat{\mathcal{L}}(\bm{\phi}, \mathcal{D}_b) \leq \widehat{\mathcal{L}}(\bm{\phi}, \mathcal{D}_i) \}. \nonumber
\end{eqnarray}
Since the algorithm employs a subgradient ascent approach, convergence to a local maximum is ensured under standard assumptions, following classical results in non-convex optimization~\cite{mohri2018foundations}. By leveraging this subgradient approach, each of the BS can iteratively refine its local estimate without requiring exact gradients, thus enabling an efficient and communication-friendly distributed estimation procedure.
\vspace{-0.2cm}
\section{Simulation Results}
\label{sec:sim_res}
To validate the performance of the proposed algorithms, datasets from \cite{lo17} are utilized. The dataset comprises tracked head movements of $50$ users while watching a catalog of $10$ HD $360^{\circ}$ YouTube videos. A $100^{\circ} \times 100^{\circ}$ FOV is considered, and to build the tiled-FOV, the equirectangular projection of each of the video frames is divided into $ N \times P$ tiles. The proposed DMTFL-based algorithm is compared with the following benchmark methods:
\begin{itemize}
    \item FedAvg Algorithm: The central BS gathers the gradients from each BS's local loss to train a global model that minimizes the overall loss across all BSs \cite{pmlr-mcmahan17a}.
    \item FedProx Algorithm: FedProx algorithm extends FedAvg by adding a proximal term to the local objective at each BS
    \cite{li2020federated}.
    \item Heuristic Caching: In this caching algorithm, it selects and caches the tiles with the highest overall global request frequency (i.e., most popular tiles across all users), regardless of individual user or BS preferences.
\end{itemize}
Fig.~\ref{fig:com} illustrates the impact of the cache size of the BS on the average cache hit. As depicted in the figure, a clear trend emerges, indicating that as the cache size of the BS increases, the average cache hit naturally increases. It is clear from Fig.~\ref{fig:com} that the proposed algorithm performs better than the baseline algorithms, demonstrating the benefit of using the proposed algorithm. For instance, the average cache hit for the algorithm is at least $55$\% higher as compared to the benchmark algorithms.
\begin{figure}[t!]
   \centering
   \begin{minipage}[b]{0.48\columnwidth}
      \centering
      \resizebox{0.9\textwidth}{0.9\textwidth}{\begin{tikzpicture}[thick,scale=1, every node/.style={scale=1.3},font=\Huge]
%
%
\definecolor{mycolor1}{rgb}{0.92900,0.69400,0.12500}%
\definecolor{mycolor2}{rgb}{0.49400,0.18400,0.55600}%
\definecolor{mycolor3}{rgb}{0.46600,0.67400,0.18800}%
\definecolor{mycolor4}{rgb}{0.30100,0.74500,0.93300}%
\definecolor{mycolor5}{rgb}{0.63500,0.07800,0.18400}%
\definecolor{mycolor6}{rgb}{1.00000,0.00000,1.00000}%
\definecolor{mycolor7}{rgb}{0.00000,0.00000,1.00000}

\begin{axis}[%
width=9.2in,
height=7.9in,
at={(1.06in,0.651in)},
scale only axis,
xmin=10,
xmax=50,
xtick={10, 20,..., 50},
xlabel style={font=\color{white!15!black}},
xlabel={\Huge{Cache size}},
ymin =  0,
ymax=0.9,
ytick={0.1,0.2,..., 0.9},
ylabel style={font=\color{white!15!black}},
ylabel={\Huge{Average cache hit}},
axis background/.style={fill=white},
xmajorgrids,
ymajorgrids,
legend style={legend cell align=left, align=left, draw=white!15!black}
]

\addplot [color=mycolor7, line width=2.0pt, mark=o, mark size=6.0pt, mark options={solid, mycolor7}]
  table[row sep=crcr]{%
10	0.18\\
20	0.335\\
30	0.492\\
40	0.64\\
50  0.78\\
};
\addlegendentry{DMTFL Algo}

\addplot [color=red, line width=2.0pt, mark=asterisk, mark size=8.0pt, mark options={solid, red}]
  table[row sep=crcr]{%
10	0.15\\
20	0.205\\
30	0.31\\
40	0.45\\
50  0.56\\
};
\addlegendentry{FedProx}

\addplot [color=mycolor1, line width=2.0pt, mark=triangle, mark size=8.0pt, mark options={solid, mycolor1}]
  table[row sep=crcr]{%
10	0.12\\
20	0.16\\
30	0.196\\
40	0.22\\
50  0.23\\
};
\addlegendentry{FedAvg}

\addplot [color=mycolor6, line width=2.0pt, mark=square, mark size=6.0pt, mark options={solid, mycolor6}]
  table[row sep=crcr]{%
10	0.09\\
20	0.14\\
30	0.16\\
40	0.187\\
50  0.2\\
};
\addlegendentry{Heuristic Caching}

\end{axis}
			\end{tikzpicture}}
   \caption{\small{Average cache hit versus cache size.}}
		\label{fig:com}
\end{minipage}
    \hfill
    \begin{minipage}[b]{0.48\columnwidth}
   \centering
  \resizebox{0.9\textwidth}{0.9\textwidth}{\begin{tikzpicture}[thick,scale=1, every node/.style={scale=1.3},font=\Huge]
%
%
\definecolor{mycolor1}{rgb}{0.92900,0.69400,0.12500}%
\definecolor{mycolor2}{rgb}{0.49400,0.18400,0.55600}%
\definecolor{mycolor3}{rgb}{0.46600,0.67400,0.18800}%
\definecolor{mycolor4}{rgb}{0.30100,0.74500,0.93300}%
\definecolor{mycolor5}{rgb}{0.63500,0.07800,0.18400}%
\definecolor{mycolor6}{rgb}{1.00000,0.00000,1.00000}%
\definecolor{mycolor7}{rgb}{0.00000,0.00000,1.00000}

\begin{axis}[%
width=9.2in,
height=7.9in,
at={(1.06in,0.651in)},
scale only axis,
xmin=5,
xmax=25,
xtick={5,10,...,25},
xlabel style={font=\color{white!15!black}},
xlabel={\Huge{Cache Size}},
ymin = 0,
ymax = 0.5,
ytick={0.0,0.1,...,0.5},
ylabel style={font=\color{white!15!black}},
ylabel={\Huge{Minimum per BS average cache hit}},
axis background/.style={fill=white},
xmajorgrids,
ymajorgrids,
legend style={legend cell align=left, align=left, draw=white!15!black}
]
\addplot [color=mycolor7, line width=2.0pt, mark=square, mark size=6.0pt, mark options={solid, mycolor7}]
  table[row sep=crcr]{%
5	0.1\\
10	0.18\\
15	0.24\\
20	0.32\\
25  0.4\\
};
\addlegendentry{DMFTL Algo}

\addplot [color=red, line width=2.0pt, mark=asterisk, mark size=8.0pt, mark options={solid, red}]
  table[row sep=crcr]{%
5	0.07\\
10	0.08\\
15	0.11\\
20	0.16\\
25  0.22\\
};
\addlegendentry{FedProx}

\addplot [color=mycolor1, line width=2.0pt, mark=triangle, mark size=8.0pt, mark options={solid, mycolor1}]
  table[row sep=crcr]{%
5	0.06\\
10	0.075\\
15	0.08\\
20	0.091\\
25  0.099\\
};
\addlegendentry{FedAvg}

\addplot [color=mycolor6, line width=2.0pt, mark=square, mark size=6.0pt, mark options={solid, mycolor6}]
  table[row sep=crcr]{%
5	0.05\\
10	0.069\\
15	0.074\\
20	0.084\\
25  0.090\\
};
\addlegendentry{Heuristic Caching}

\end{axis}
		\end{tikzpicture}}
\caption{\small{Minimum average cache hit versus cache size.}}
	\label{fig:fair}
\end{minipage}
\end{figure}

Fig.~\ref{fig:fair} illustrates the minimum per-BS cache hit rate, which serves as a fairness indicator by highlighting the worst-served BS in the network. Unlike average or aggregate metrics, this measure directly captures the performance experienced by the most disadvantaged BS. A higher minimum hit rate indicates that the algorithm is effectively avoiding bias toward data-rich or majority BSs, aligning with the FL fairness objective of providing balanced and equitable performance across all BSs. As seen in the Fig.~\ref{fig:fair}, the proposed DMTFL algorithm achieves a significantly higher minimum per-BS hit rate compared to baseline methods, demonstrating its superior fairness properties.
\section{Conclusion}
\label{sec:concl}
\noindent This paper tackles the critical challenge of VR content caching in highly dynamic and adversarial environments, leveraging the power of the FL framework. By explicitly addressing the challenges of statistical heterogeneity and fairness across BSs, the proposed approach enables personalized caching strategies tailored to each user's FOV demands. Furthermore, we established rigorous theoretical foundations through Rademacher complexity analysis and derived a PAC bound on the loss given by Theorem \ref{thm:mainresult_cachingvr}, providing strong generalization guarantees. The numerical results demonstrate the superior performance of our proposed algorithm compared to existing baseline methods, highlighting its effectiveness in handling future demands of dynamic VR content delivery. While this paper demonstrates the potential of the proposed DMTFL algorithm, several important aspects remain open for future investigation. Evaluating communication and computation overhead in practical MEC environments, especially as the number of BSs increases, is also a crucial direction. Furthermore, the proposed framework can also be extended to emerging music metaverse applications, where immersive VR environments are further enriched with dynamic music content. Finally, extending the framework to account for dynamic user mobility and BS associations will be considered in future work.
\section*{Acknowledgment}
\noindent We acknowledge the support of the MUSMET project funded by the EIC Pathfinder Open scheme of the European Commission (grant agreement no. 101184379).
\appendices
\section{Proof of Theorem \ref{thm:mainresult_cachingvr}} \label{appendix:proof_main_res_cachingvr}
\noindent Consider $\mathcal{D}_b = \{(x_{b,1}, y_{b,1}), (x_{b,2}, y_{b,2}), \ldots, (x_{b,m_b}, y_{b,m_b})\}$, which represents the set of data samples at the BS $b$. Each BS $b$ is assumed to have caching strategy $\bm{\phi}_b \in \mathbb{R}^F$, for $b = 1, 2, \ldots, B$. Let ${\Phi} = \{\bm{\phi}_1, \bm{\phi}_2, \ldots, \bm{\phi}_B\} \subset \mathbb{R}^{F \times B}$ denote the collection of all caching strategies across the BSs. The weights $\bm{\phi}_b$ at the BS $b$ are optimized using the loss function
\vspace{-0.2cm}
\begin{eqnarray}
L_b(\bm{\phi}_b) := \mathbb{E}_{(x_b, y_b) \sim \mathcal{D}_b}\bigl[l(h_{\bm{\phi}_b}(x_b), y_b)\bigr]. \nonumber
\end{eqnarray}
The weighted sum of the average losses across all the BS in the network is given by
\vspace{-0.2cm}
\begin{eqnarray}
\hspace{-0.2cm}\theta_{\Phi,\bm{w}} \hspace{-0.2cm}&=& \hspace{
-0.2cm
} \sum_{b=1}^B w_b \mathcal{L}_b(\bm{\phi}_b) \nonumber \\
\vspace{-0.18cm}
\hspace{-0.3cm}&\stackrel{(a)}{=}&\hspace{-0.3cm}  \sum_{b,i=1}^B  w_b \alpha_{b,i}\mathcal{L}_i(\bm{\phi}_b) + \sum_{b,i =1}^Bw_b\alpha_{b,i} \big[\mathcal{L}_b(\bm{\phi}_b) - \mathcal{L}_i(\bm{\phi}_b)\big],\nonumber
\end{eqnarray}
\vspace{-0.38cm}
\begin{eqnarray}
&\stackrel{(b)}{=}& \theta_{\Phi, \mathbf{w}, \bm{\alpha}} + \sum_{b=1}^B\sum_{i=1}^B w_b\alpha_{b,i} \big[\mathcal{L}_b(\bm{\phi}_b) - \mathcal{L}_i(\bm{\phi}_b)\big],\nonumber\\
&\stackrel{(c)}{\leq}& \theta_{\Phi, \mathbf{w}, \bm{\alpha}} + \sum_{b=1}^B\sum_{i=1}^Bw_b \alpha_{b,i}v_{b,i}\nonumber
\end{eqnarray}
where $(a)$ follows from adding and subtracting the term $\sum_{b=1}^B\sum_{i=1}^Bw_b\alpha_{b,i}\mathcal{L}_i(\phi_b)$, and using the fact that $\sum_{b=1}^B w_b = 1$, and $\sum_{i=1}^B \alpha_{b,i}^B = 1$, $(b)$ follows from using the definition of $\theta_{\Phi, \mathbf{w}, \bm{\alpha} }=  \sum_{b=1}^B \sum_{i=1}^B  w_b \alpha_{b,i}\mathcal{L}_i(\phi_b)$, and $(c)$ follows from the definition of discrepancy between two BSs $b$ and $i$, i.e., $v_{b,i} = \sup_{\bm{\phi}}| \mathcal{L}_i(\bm{\phi}) - \mathcal{L}_j(\bm{\phi})|$.
Thus, the above can be upper bounded as follows:
\vspace{-0.35cm}
\begin{eqnarray}
    \theta_{\Phi,\mathbf{w}} \leq \theta_{\Phi, \mathbf{w}, \bm{\alpha}} + \sum_{b=1}^B\sum_{i=1}^Bw_b \alpha_{b,i}v_{b,i}.
        \label{eq:first_bound}
\end{eqnarray}   
Note that $\widehat{\theta}_{\Phi, \mathbf{w}, \bm{\alpha}} = \sum_{b=1}^B w_b\sum_{i=1}^B\alpha_{b,i}\mathcal{\hat{L}}_b(\bm{\phi}_{i}, \mathcal{D}_b)$ is an estimate of the term $\theta_{\Phi, \mathbf{w}, \bm{\alpha}}$. Let $\Upsilon(\mathcal{D}) = \sup_{\bm{\phi} \in \Phi}\big( \theta_{\Phi, \mathbf{w}, \bm{\alpha}} - \widehat{\theta}_{\Phi, \mathbf{w}, \bm{\alpha}}(\mathcal{D})\big)$. Let us consider two sets of samples, $\mathcal{D}'_b = \{\mathcal{D}'_1, \ldots, \mathcal{D}'_B\}$ and $\mathcal{D}_b = \{\mathcal{D}_1, \ldots, \mathcal{D}_B\}$, which differ in exactly one data point. Specifically, let this differing element be $(x'_{b,i}, y'_{b,i}) \in \mathcal{D}'_b$ and $(x_{b,i}, y_{b,i}) \in \mathcal{D}_b$, where $b \in \{1, \ldots, B\}$ and $i \in \{1, \ldots, m_b\}$. To apply McDiarmid's inequality~\cite{mohri2018foundations}, it is necessary to bound the following difference:
\begin{eqnarray}
 \Upsilon(\mathcal{D}') - \Upsilon(\mathcal{D}) &=& \sup_{\bm{\phi} \in \Phi} \big( \theta_{\Phi, \mathbf{w}, \bm{\alpha}} - \widehat{\theta}_{\Phi, \mathbf{w}, \bm{\alpha}}(\mathcal{D}')\big) \nonumber \\
        &-&\sup_{\bm{\phi} \in \Phi} 
        \big( \theta_{\Phi, \mathbf{w}, \bm{\alpha}} - \widehat{\theta}_{\Phi, \mathbf{w}, \bm{\alpha}}(\mathcal{D})\big) \nonumber
\end{eqnarray}
\vspace{-0.5cm}
\begin{eqnarray}
& \stackrel{(a)}{\leq}& \sup_{\bm{\phi} \in \Phi} 
\big[\big( \theta_{\Phi, \mathbf{w}, \bm{\alpha}} - \widehat{\theta}_{\Phi, \mathbf{w}, \bm{\alpha}}(\mathcal{D}')\big) - \big( \theta_{\Phi, \mathbf{w}, \bm{\alpha}} - \widehat{\theta}_{\Phi, \mathbf{w}, \bm{\alpha}}(\mathcal{D})\big) \big] \nonumber \\
        &\leq& \sup_{\bm{\phi} \in \Phi} 
        \big( \widehat{\theta}_{\Phi, \mathbf{w}, \bm{\alpha}}(\mathcal{D}) - \widehat{\theta}_{\Phi, \mathbf{w}, \bm{\alpha}}(\mathcal{D})'\big) \nonumber\\
        & \leq & \sup_{\bm{\phi} \in \Phi} \bigg( \sum_{b,i = 1}^B w_b\alpha_{b,i}\big[\mathcal{\hat{L}}_b(\bm{\phi}_{i}, \mathcal{D}_b) - \mathcal{\hat{L}}_b(\bm{\phi}_{i}, \mathcal{D}'_b) \big] \bigg) \nonumber \\
        & \stackrel{(b)}\leq& \sum_{b,i=1}^B \frac{w_b}{m_i} \alpha_{b,i}H, \nonumber
\end{eqnarray}
where $(a)$ follows from the properties of the supremum, while $(b)$ follows from the fact that the loss is assumed to be bounded. Applying McDiarmid’s inequality for any $\delta > 0$ and $\bm{\phi} \in \Phi$,  it can be shown that the following holds with probability at least $1 - \delta$,
\begin{eqnarray}
    \theta_{\Phi, \mathbf{w}} &\leq &\widehat{\theta}_{\Phi, \mathbf{w}, \bm{\alpha}}(\bm{\mathcal{D}}) + \mathbb{E}_{\mathcal{D}}[\Upsilon(\mathcal{D})] \nonumber \\
    && + H\sqrt{\frac{1}{2} \sum_{i=1}^B\big(\sum_{b=1}^B \frac{w_b}{m_i} \alpha_{b,i}\big)^2 \log(\frac{1}{\delta})}.\nonumber
\end{eqnarray}
Therefore, by the union over $\Lambda_{\epsilon}$, and by definition of $\Lambda_{\epsilon}$, for any $\mathbf{w} \in \Lambda$, there exists $\mathbf{w}' \in \Lambda_{\epsilon}$ such that $\widehat{\theta}_{\Phi, \mathbf{w}, \bm{\alpha}}(\bm{\mathcal{D}}) \leq \widehat{\theta}_{\Phi, \mathbf{w}, \bm{\alpha}}(\bm{\mathcal{D}'}) + HB\epsilon$. In light of this, with probability of $1 - \delta$, the following holds
\begin{eqnarray}
\hspace{-0.6cm}    \theta_{\Phi, \mathbf{w}} &\leq &\widehat{\theta}_{\Phi, \mathbf{w}, \bm{\alpha}}(\bm{\mathcal{D}}) + \mathbb{E}_{\mathcal{D}}[\Upsilon(\mathcal{D})] + HB\epsilon \nonumber\\
    &&+ H\sqrt{\frac{1}{2} \sum_{i=1}^B\big(\sum_{b=1}^B \frac{w_b}{m_i} \alpha_{b,i}\big)^2 \log(\frac{|\Lambda_{\epsilon}|}{\delta})}.
    \label{eq:second_bound}
\end{eqnarray}
For the term $\mathbb{E}_{\mathcal{D}}[\Upsilon(\mathcal{D})]$, we have the following
\begin{eqnarray}
\hspace{-0.6cm}&&\mathbb{E}_{\mathcal{D}}[\Upsilon(\mathcal{D})] = \mathop{\mathbb{E}}\limits_{\mathcal{D}}
\bigg[    \sup_{\bm{\phi} \in \Phi} \big( \theta_{\Phi, \mathbf{w}, \bm{\alpha}} - \widehat{\theta}_{\Phi, \mathbf{w}, \bm{\alpha}}(\mathcal{D}')\big) \bigg] \nonumber\\
 \hspace{-0.6cm}     && = \mathop{\mathbb{E}}\limits_{\mathcal{D}}
\bigg[    \sup_{\bm{\phi} \in \Phi} \mathbb{E}_{\mathcal{D}'}\big( \widehat{\theta}_{\Phi, \mathbf{w}, \bm{\alpha}}(\mathcal{D}') - \widehat{\theta}_{\Phi, \mathbf{w}, \bm{\alpha}}(\mathcal{D})\big) \bigg] \nonumber \\
 \hspace{-0.6cm}     && \stackrel{(a)} \leq \mathop{\mathbb{E}}\limits_{\mathcal{D}, \mathcal{D}'}
\bigg[    \sup_{\bm{\phi} \in \Phi} \big( \widehat{\theta}_{\Phi, \mathbf{w}, \bm{\alpha}}(\mathcal{D}') - \widehat{\theta}_{\Phi, \mathbf{w}, \bm{\alpha}}(\mathcal{D})\big) \bigg]\nonumber\\
 \hspace{-0.6cm}   && \leq \mathop{\mathbb{E}}\limits_{\mathcal{D}, \mathcal{D}'}
\bigg[    \sup_{\bm{\phi} \in \Phi} \bigg( \sum_{b,i =1}^B \sum_{j=1}^{m_b} \frac{w_b\alpha_{b,i}}{m_j}[\ell ( h_{\bm{\phi}}(x_{b,i}'), y_{b,i}') \nonumber\\
  \hspace{-0.6cm}  && -  \ell  (h_{\bm{\phi}}(x_{b,i}), y_{b,i}) ]\bigg) \bigg]\nonumber \\
 \hspace{-0.6cm}   && \leq \mathop{\mathbb{E}}\limits_{\mathcal{D}, \mathcal{D}', \sigma}
\bigg[    \sup_{\bm{\phi} \in \Phi} \bigg( \sum_{b,i =1}^B \sum_{j=1}^{m_b} \frac{\sigma_{bi,j}w_b\alpha_{b,i}}{m_j}[\ell ( h_{\bm{\phi}}(x_{b,i}'), y_{b,i}') \nonumber\\
 \hspace{-0.6cm}   && -  \ell  (h_{\bm{\phi}}(x_{b,i}), y_{b,i}) ]\bigg) \bigg] \nonumber\\
\hspace{-0.6cm}    &&\stackrel{(b)}\leq  \mathop{\mathbb{E}}\limits_{\mathcal{D}', \sigma}
\bigg[    \sup_{\bm{\phi} \in \Phi} \bigg( \sum_{b,i =1}^B \sum_{j=1}^{m_b} \frac{\sigma_{bi,j}w_b\alpha_{b,i}}{m_j}[\ell ( h_{\bm{\phi}}(x_{b,i}'), y_{b,i}')]\bigg) \nonumber\\
 \hspace{-0.6cm}   && + \mathop{\mathbb{E}}\limits_{\mathcal{D}, \sigma}
\bigg[ \sup_{\bm{\phi} \in \Phi} \bigg( \sum_{b,i =1}^B \sum_{j=1}^{m_b} \frac{\sigma_{bi,j}w_b\alpha_{b,i}}{m_j}[- \ell  (h_{\bm{\phi}}(x_{b,i}), y_{b,i}) ]\bigg) \bigg] \nonumber\\
\hspace{-0.6cm}    && \leq 2\mathop{\mathbb{E}}\limits_{\mathcal{D}, \sigma}
\bigg[ \sup_{\bm{\phi} \in \Phi} \bigg( \sum_{b,i =1}^B \sum_{j=1}^{m_b} \frac{\sigma_{bi,j}w_b\alpha_{b,i}}{m_j}[\ell  (h_{\bm{\phi}}(x_{b,i}), y_{b,i}) ]\bigg) \bigg]\nonumber\\
\hspace{-0.6cm}    &&\leq 2 \mathcal{R}_{\Lambda}(\Phi) \nonumber 
\end{eqnarray}
where $(a)$ follows from the Jensen's inequality, and $(b)$ follows from the standard Rademacher complexity upper bounds and from the definition of minimax weighted Rademacher complexity, and $\sigma_{bi,j}$
is the Rademacher random variable.
Using the above result in~\ref{eq:second_bound}, we get the following 
\begin{eqnarray}
\hspace{-0.6cm}    \theta_{\Phi, \mathbf{w}} &\leq &\widehat{\theta}_{\Phi, \mathbf{w}, \bm{\alpha}}(\bm{\mathcal{D}}) + 2 \mathcal{R}_{\Lambda}(\Phi) + HB\epsilon
    \nonumber\\
    &&+ H\sqrt{\frac{1}{2} \sum_{i=1}^B\big(\sum_{b=1}^B \frac{w_b}{m_i} \alpha_{b,i}\big)^2 \log(\frac{|\Lambda_{\epsilon}|}{\delta})}.
    \label{eq:third_bound}
\end{eqnarray}
Next, substituting~\eqref{eq:third_bound} in~\eqref{eq:first_bound}, we get the following:
\begin{eqnarray}
   \theta_{\Phi, \mathbf{w}}\hspace{-0.2cm} &\leq & \hspace{-0.3cm}\widehat{\theta}_{\Phi, \mathbf{w}, \bm{\alpha}}(\bm{\mathcal{D}}) + 2 \mathcal{R}_{\Lambda}(\Phi)\hspace{-0.12cm} + HB\epsilon + \sum_{b=1}^B\sum_{i=1}^Bw_b \alpha_{b,i}v_{b,i}
    \nonumber\\
    && \hspace{-0.2cm}+ H\sqrt{\frac{1}{2} \sum_{i=1}^B\big(\sum_{b=1}^B \frac{w_b}{m_i} \alpha_{b,i}\big)^2 \log(\frac{|\Lambda_{\epsilon}|}{\delta})}.
    \label{eq:last_bound} 
\end{eqnarray}
This proves the theorem.
\vspace{-0.1cm}
\bibliographystyle{IEEEtran}
\bibliography{strings}
\end{document}